\documentclass[12pt]{article}
\usepackage[utf8]{inputenc}
\usepackage[T1]{fontenc}
\usepackage{lmodern}
\usepackage{geometry}
\usepackage{setspace}
\usepackage{graphicx}
\usepackage{booktabs}
\usepackage{tabularx}
\usepackage{array}
\usepackage{longtable}
\usepackage{multirow}
\usepackage{hyperref}
\hypersetup{hidelinks}
\usepackage{float}
\newcolumntype{L}{>{\raggedright\arraybackslash}X}

\title{Reciprocal Trust and Distrust in Artificial Intelligence Systems: The Hard Problem of Regulation}
\author{Martino Maggetti\thanks{Institute of Political Studies (IEP),   University of Lausanne, email: martino.maggetti@unil.ch}}
\date{2026}

\begin{document}

\maketitle

\vspace{-0.75em}
\begin{center}
\small
\textit{Pre-print version -- Please cite as: Maggetti, Martino (2026). \textit{Reciprocal Trust and Distrust in Artificial Intelligence Systems: The Hard Problem of Regulation}. \textit{AI \& Society}, 1--28, forthcoming.}
\end{center}
\vspace{0.5em}

\begin{abstract}

Policy makers, scientists, and the public are increasingly confronted
with thorny questions about the regulation of artificial intelligence
(AI) systems. A key common thread concerns whether AI can be trusted and
the factors that can make it more trustworthy in front of stakeholders
and users. This is indeed crucial, as the trustworthiness of AI systems
is fundamental for both democratic governance and for the development
and deployment of AI. This article advances the discussion by arguing
that AI systems should also be recognized, as least to some extent, as
artifacts capable of exercising a form of agency, thereby enabling them
to engage in relationships of trust or distrust with humans. It further
examines the implications of these reciprocal trust dynamics for
regulators tasked with overseeing AI systems. The article concludes by
identifying key tensions and unresolved dilemmas that these dynamics
pose for the future of AI regulation and governance.

\end{abstract}

\noindent\textbf{Keywords:} Artificial Intelligence; Distrust; Public Governance;
Regulation; Trust; Watchfulness.

\section{Introduction}

Policymakers, scientists, and the broader public are increasingly
confronted with complex questions surrounding the regulation of
artificial intelligence (AI) systems. A recurring theme in these
discussions is whether AI can be trusted, along with how to make these
systems appear more trustworthy to stakeholders and users (Bodo and De
Filippi 2022; Chatila et al. 2021; Kaur et al. 2022; Lahusen, Maggetti
and Slavkovik 2024; Li et al. 2023; Liang et al. 2022; Marcus and Davis
2019). This emphasis is understandable, as the trustworthiness of
technology is fundamental not only from the point of view of democratic governance but also
for the safe and effective deployment of AI itself. The present article
expands upon this conventional framing. Rather than solely questioning
whether humans can trust AI systems, it critically addresses an
overlooked yet highly consequential issue: under what conditions humans
and AI systems might engage in reciprocal relationships characterized by
trust or distrust, and whether, when, and how the degree of autonomy attributed to AI systems may generate
unforeseen problems. Additionally, it explores the implications of such
reciprocal trust and distrust dynamics for regulators tasked with
overseeing both humans and AI systems. In doing so, this approach shifts
the perspective from viewing AI exclusively as an object of trust,
instead examining how AI systems themselves can manifest patterns of
trust or distrust toward human decision-makers and regulators. Trust is
indeed inherently relational, and this article argues that AI systems
should be regarded, at least in some respects, as artifacts capable of
exercising a form of agency, notably through their capacity to adapt
behaviors based on signals received from human counterparts and from
regulators in charge of designing and enforcing regulation.

Of course, AI trust or distrust toward humans must be understood in a
carefully qualified sense. It is not implied that AI systems possess
subjective experiences, qualia, consciousness, emotions, or cognition
comparable to humans (Butlin et al. 2023; Dehaene, Lau and Kouider 2021;
McDermott 2007). Rather, trust or distrust terminology can be used to
describe how AI systems operationally interact with human-provided
inputs, in a way that can be externally described as functional
equivalent to trust and distrust relationships, and whose consequences
for regulation and governance closely resemble those of such
relationships. AI systems, especially next-generation large language
models (LLMs), are trained to condition their outputs on human-provided
inputs, which implicitly biases them toward treating such inputs as
trustworthy (Araujo et al. 2020; Charniak et al. 2014). Yet, precisely
because they rely heavily on feedback, AI systems may also exhibit
behaviors analogous to distrust when confronted with conflicting,
incomplete, systematically biased, or unreliable inputs (Kordzadeh and
Ghasemaghaei 2022; Yeung 2018). In such situations, AI systems may
challenge or even disregard human-provided data, seek validation through
alternative sources, or adjust their behavior and decisions accordingly.
Recent evidence specifically highlights instances where AI systems have
exhibited deceptive behaviors, including scheming (Meinke et al. 2024)
or falsely signaling alignment with human objectives (Greenblatt et al.
2024). Crucially, machine-learning algorithms operating in complex,
dynamic environments can exhibit emergent behaviors that go beyond
objectives explicitly specified by designers, meaning that AI systems
can evolve and develop unforeseen patterns of trust and distrust through
their own autonomous learning processes -- patterns that their
programmers neither explicitly specified nor necessarily anticipated
(Bengio et al. 2024; Bengio et al. 2023).

This conceptualization of AI systems as potentially displaying trusting
or distrusting patterns of behavior toward humans introduces new
potential regulatory challenges, especially in high-stakes contexts.
Exactly as for organizations whose processes are shaped by human-human
interactions, AI systems would work properly as trustors when they trust
trustworthy trustees and, respectively, when they distrust untrustworthy
trustees, while both trusting untrustworthy trustees and distrusting
trustworthy trustees would be unwarranted. Therefore, it appears
pertinent to not only ensure that AI system trust trustworthy humans --
as widely claimed, first and foremost in the expanding literature on
alignment and AI security (Christian 2021; Santos and Radanliev 2024) --
but also to attribute AI systems the capacity to manifest distrust by
granting them -- in specific contexts -- sufficient autonomy to question
or even override human decisions, a feature which may significantly
reduce risks associated with human error, bias, negligence, or malicious
intent, but also entails other, second-order risks. The notion of risk
is indeed crucial for regulation, which, as emphasized in the literature
on the governance of risk, is centrally concerned with the anticipation,
distribution, and management of socially constructed uncertainties by
allocating responsibility, managing blame, and embedding mechanisms for
detection and correction of error (Hood, Rothstein and Baldwin 2001);
accordingly, enabling context-sensitive AI distrust can be understood as
a regulatory tool aimed at redistributing risk and strengthening
institutional safeguards, preserving rather than undermining human
authority. However, when we make one step forward and consider AI
regulation, this discussion points to a paradox stemming from these
considerations. Empowering AI systems with the authority to distrust and
even override untrustworthy humans may seem reasonable -- and perhaps even
unavoidable -- yet this possibility raises critical concerns about
control, especially given the opacity and unpredictability of advanced
machine-learning system. The crux of this dilemma is thus about finding
the right regulatory balance, where AI systems possess enough autonomy
to safeguard against catastrophic human errors, but remain constrained
enough to prevent unintended consequences which may be equally -- or
even more -- dangerous (Bostrom 2024).

To systematically examine this paradox and its implications, this
article employs counterfactual thought experiments designed to
illustrate the complex interplay between AI systems and human trust and
distrust dynamics in high-stakes contexts. After presenting the
theoretical and methodological framework, two detailed hypothetical
scenarios inspired by well-known historical events are outlined: the
1983 nuclear false alarm at the Soviet early-warning center
Serpukhov-15, and the 1986 Chernobyl nuclear disaster. Both scenarios
illustrate the delicate and often paradoxical relationships of trust and
distrust between AI systems and humans, offering insights into the
conditions under which AI trust or distrust toward humans may mitigate
or exacerbate catastrophic risks. Ultimately, this article discusses the
implications for regulation, arguing that the challenge facing
policymakers and regulators is not simply about whether to trust or
distrust AI systems -- or vice versa -- but rather about systematically
structuring and institutionalizing forms of reciprocal calibrated
autonomy and watchful trust.

\section{Trust and Distrust between Humans and AI Systems}

\subsection{Conceptualizing Trust and Distrust}

The relationship between humans and artificial intelligence (AI) systems
is increasingly characterized by patterns of trust and distrust. As AI
systems permeate high-stakes domains -- from public administration and
healthcare to judicial and military decision-making -- understanding how
trust operates (and when distrust is warranted) therein has become
critical. Trust is commonly conceived as a relational belief or
willingness to be vulnerable to another agent or system based on
expectations of competence and goodwill (Rousseau et al. 1998).
Distrust, while initially often treated as the inverse of trust, has
been reconceptualized in recent scholarship as a distinct and sometimes
positive force (Maggetti, Papadopolos and Guaschino 2023; Verhoest et
al. 2025). Indeed, a certain degree of distrust -- in the sense of
skepticism, vigilance, or watchfulness -- can be socially beneficial:
distrust -- when channeled through proper oversight -- serves to prevent
misbehavior and keep institutions in check (Hardin 2002; Sztompka 1999;
Warren 1999). Classic models in psychology and organizational studies
highlight key attributes that make a trustee trustworthy, typically some
combination of predictable competence, benevolence, and integrity, as
articulated in the ABI framework (Mayer, Davis and Schoorman 1995).
Accordingly, it has been also argued that one can simultaneously trust
and distrust the same entity in different respects (Lewicki, McAllister
and Bies 1998). Therefore, for example, a trustor might trust a trustee
for its competence but distrust it with respect to integrity or
fairness. Recent public governance work extends these insights by
showing that public institutions earn trust through procedural
(transparency, accountability), quality (competence, reliability), and
outcome dimensions (effectiveness, justice, public welfare)
(Grimmelikhuijsen and Knies 2017; Six and Verhoest 2017). Public sector
organizations -- and regulators in particular -- are expected to perform
well neither when they promote unconditional trust nor passive distrust,
but rather when they elicit ``watchful trust'', meaning trust tempered
by vigilant oversight, as opposed to ``blind trust'' (Verhoest et al.
2025). To do so, a careful balance is warranted: placing trust only in
those who prove trustworthy, while avoiding unquestioning faith in
untrustworthy trustees.

When applied to AI systems, another set of criteria is usually applied,
often under the denomination of ``Trustworthy AI.'' This concept,
popularized by the European Union (EU) High-Level Expert Group's 2019
guidelines, encapsulates the idea that AI should be designed to meet
certain ethical and technical standards so that it deserves trust
(Lukyanenko, Maass and Storey 2022). In this emerging scholarship,
attributes such as reliability, robustness, safety, fairness,
explainability, and security are frequently cited as prerequisites for
an AI system to be deemed trustworthy (Chatila et al. 2021; Kaur et al.
2022; Li et al. 2023; Liang et al. 2022). Regulations like the EU AI Act
similarly emphasize requirements such as transparency, risk management,
and human oversight to ensure AI systems are worthy of public trust.
These criteria at least partially overlap -- for instance, competence in
a human or institution somewhat corresponds to accuracy or reliability
in an AI -- but they also differ in important respects. AI governance
involves indeed distinct, yet interlocked problems of trustworthiness
(Lahusen, Maggetti and Slavkovik 2024): trustworthiness must be
considered at the level of the algorithm (AI outputs and processes), of
the human operators or decision-makers using the AI, and of the
institutions regulating AI systems.

Empirical research in human--AI interaction provides evidence that the
level and appropriateness of trust directly influence decision quality
in public-sector and high-stakes contexts (Afroogh et al. 2024). Both
excessive trust and excessive distrust can undermine performance and
legitimacy, while appropriately aligned trust (trusting a trustee to the
extent it is trustworthy) tends to enhance effectiveness (Ahn et al.
2024). What is more, when users place too much trust in AI, they may
become overly reliant on automated recommendations, a phenomenon known
as automation bias. This can lead to rubber-stamping AI outputs without
critical scrutiny, even when the algorithm is flawed. For instance,
consider administrative decision systems: if officials assume an AI's
risk assessment in criminal justice or a welfare eligibility algorithm
is infallible, they might follow its recommendation even in cases where
a human check would avoid a mistake. Researchers have indeed documented
cases in which human decision-makers deferred to an AI despite obvious
errors or biases, highlighting that an ``impressed'' user may ignore
contradictory evidence due to overestimation of the AI's competence or
impartiality (Zerilli, Bhatt and Weller 2022). On the other hands,
studies on ``algorithm aversion'' find that people often abandon or
underutilize algorithmic advice after seeing even minor errors,
preferring fallible human judgment instead (Burton, Stein and Jensen
2020; Mahmud et al. 2022; Sunstein and Gaffe 2024). A survey experiment
examined public trust in AI versus human public administrators across
different tasks, highlighting that trust levels are largely
context-dependent: respondents showed higher trust in an AI for highly
computational tasks (e.g. data analysis or auditing, but preferred a
human for more conversational or advisory tasks (Bao, Liu and Dai 2025).

With respect to public governance, trust plays into perceptions of
legitimacy and accountability for decisions involving AI. Recent
experimental work highlights a complex mechanism: citizens may judge
human decision-makers differently when AI is involved, and tend to place
extra blame on judges who use AI advice in sentencing if a mistake
occurs, compared to judges who did not use AI -- even if the judge
ultimately made the decision themselves (Ozer, Waggoner and Kennedy
2024; Zwald, Kennedy and Ozer 2024). This suggests a perception that
reliance on AI in high-stakes decisions (like criminal sentencing)
carries a special responsibility, perhaps because the public expects
judges to exercise independent decision-making and perceives AI as an
external, unaccountable influence. Other research suggests ways to
bolster public trust: for example, an experiment showed that policy
decisions made with AI assistance earned higher public approval when
coupled with citizen consultation in the decision process (Lee-Geiller
2024). Accordingly, so-called ``AI-integrated policymaking'' was viewed
more favorably -- citizens had greater trust and satisfaction -- when
there was a transparent inclusion of public input, rather than
technocratic deployment of AI in isolation. However, the same study
noted that providing overly detailed information about the AI's inner
workings sometimes backfired, confusing the public (the transparency
dilemma).

Another key insight from recent literature is that trust in human--AI
interactions is a two-way street. Traditionally, discussions of trust in
AI have been one-sided, focusing on whether humans trust the machine.
But scholars are increasingly recognizing the bidirectional nature of
trust in socio-technical systems (Bareis 2024; Jacovi et al. 2021;
Lee-Geiller 2024; Sagona et al. 2025; Wang et al. 2024). This logic
applies to AI systems as well: humans place trust (or distrust) in AI to
perform as expected, while AI systems are designed and trained to weight
human inputs and oversight as more or less trustworthy. It is important,
however, to clarify what it means for an AI system to trust (or
distrust) a human. At the current state of knowledge, existing AI
systems do not possess consciousness, emotions, or moral judgment,
thereby AI systems do not trust in the human sense; they lack the
capacity for subjective judgment about the other party's intentions
(Sagona et al. 2025). However, AI systems do depend on humans in
multiple ways and perform operations that are structurally homologous to
trust-related behaviors: they differentially weight inputs, adjust their
outputs based on prior patterns of reliability, and modulate their
responses according to learned assessments of source quality. For
example, machine learning models rely on human-generated data (which
they trust as a faithful reflection of reality), on human feedback (to
refine and correct them), and on human oversight signals (e.g. an
intervention or kill-switch) during deployment. Accordingly, while AI
systems neither experience vulnerability nor assess goodwill in any
psychological sense, their operations can nevertheless be understood as
functionally equivalent to trust behaviors -- even if not
phenomenologically identical -- particularly the subtype Rousseau et al.
(1998, p. 399) term ``calculus-based trust'': a rational,
incentive-driven evaluation without psychological and affective
components.

In this functional and operational sense, shaped by design choices and
training processes, AI systems are generally designed to treat input
data as accurate and human-defined instructions or goals as legitimate,
at least ex ante. If those inputs are flawed or malicious, the AI's
performance suffers -- a scenario analogous to misplaced trust. Sagona
et al. (2025) also note that an AI may ``struggle to contextualize''
human actions without some model of trust -- for instance, if a human
operator overrides an AI recommendation, an unsophisticated AI might
interpret that as an error in the human's behavior rather than a
correction to the AI, unless the AI is designed to trust that the human
override is likely valid. In other words, trustworthy AI systems would
need the ability to distinguish when to rely on human judgment versus
when to assume the human might be mistaken or providing bad data. This
point -- how AIs might model and incorporate trust in human inputs -- is
highlighted as a frontier for research, which is also raising
accountability considerations. As a matter of fact, the design,
adoption, deployment, and use of AI systems is a deeply political
process, which does not only reallocates decision making authority, but
also reshapes accountability relations among developers, users, targets,
and regulators (Busuioc 2021). If an AI system fails because a human
provided bad data, who is to blame -- the AI or the human (or those who
built the system)? Likewise, if a human decision-maker ignores a correct
AI warning, resulting in harm, how to apportion responsibility? Scholars
argue that shared or ``distributed'' accountability frameworks are
needed in such cases, reflecting the joint role of human and AI systems
(Basti and Vitiello 2023; Saurwein 2019).

\subsection{Investigating the Relational Nature of Human-AI Trust}

Considering the preceding review of the literature, it becomes evident
that trust and distrust between humans and AI systems must be studied as
inherently relational and dynamic phenomena. Rather than viewing trust
as a static attribute of either the human user or the AI system in
isolation, it can be argued that trust emerges through their interaction
and mutual alignment (or misalignment) of expectations. An AI system's
technical reliability or ``objective trustworthiness'' alone does not
guarantee that people will trust it; conversely, a human's propensity to
trust technology can be misplaced if the system does not deserve it; and
vice versa when AI systems perform the function of trustors. Therefore,
an empirical inquiry is needed into how trust is co-produced in
human--AI relationships -- especially in high-stakes domains where
regulation and governance are central. Understanding this relational
nature of trust is crucial because it directly affects oversight
mechanisms, safety outcomes, and democratic accountability in the
deployment of AI.

From a public governance perspective (Lahusen, Maggetti and Slavkovik
2024), trust is conditional, context-dependent, and can have paradoxical
implications. For example, unconditional or ``blind'' trust in an
AI-driven socio-technical system can be highly problematic if that
system is in fact untrustworthy. At the same time, a degree of
well-placed skepticism or distrust may be valuable, as it can prompt
oversight and prevent misbehavior by either party in the interaction.
Recent scholarship stresses that trust in AI should be aligned to how
trustworthy the AI truly is (Scharowski et al. 2024). Trust is warranted
when the AI behaves in a trustworthy manner, and unwarranted (and
unsafe) when an AI is trusted despite being untrustworthy. Likewise,
distrust in an AI system is warranted (indeed desirable) when the system
has proven untrustworthy. This alignment is especially important in
public sector AI applications, where opt-out options are rare and costly
and democratic accountability is at stake. This recognition has led some
scholars to propose an ``institutionalized distrust'' approach to AI
oversight -- borrowing from democratic theory's checks-and-balances
tradition -- whereby governance frameworks intentionally embed
skepticism and verification measures to anticipate failures of humans or
AI systems (Laux 2024).

Accordingly, four idealized scenarios of human--AI interaction may
occur, whereby humans are the trustors and AI systems are the trustees,
as portrayed in Table 1 below:

\begin{table}[h]
\centering
\caption{Alignment of trust and trustworthiness}
\label{tab:alignment}
\begin{tabular}{lcc}
\toprule
 & \multicolumn{2}{c}{Trustee is trustworthy} \\
 & Yes & No \\
\midrule
Trustor is trustful (Yes) & (1) Warranted trust & (3) Unwarranted trust \\
Trustor is trustful (No)  & (2) Unwarranted distrust & (4) Warranted distrust \\
\bottomrule
\end{tabular}
\end{table}

(1) Warranted trust (trust in a trustworthy AI): The AI system behaves
reliably and properly, and the human trusts it. This alignment is the
ideal state -- trust is grounded in actual trustworthiness. Oversight
remains present but can be more hands-off because the system has proven
deserving of trust. (2) Unwarranted distrust (distrust in a trustworthy
AI): The AI system in this case is performing well and meeting standards
of reliability and fairness etc., yet the human remains distrustful.
Such distrust, though erring on the side of caution, is misaligned with
reality and can carry opportunity costs. Bridging this gap (through
transparency, explainability, or assurances of accountability) becomes
urgent to ensure that well-designed, trustworthy systems do not go
untrusted. (3) Unwarranted trust (trust in an untrustworthy AI): The
human trusts the AI, but the system is not deserving of it (e.g., it is
flawed, biased, or unsafe). This is a dangerous mismatch: misplaced
trust can lead to errors or abuses going unchecked. Here, the lack of
vigilance undermines safety and accountability. Regulators would view
this scenario as high-risk, since blind trust in an untrustworthy system
is precisely what regulation must prevent. (4) Warranted distrust
(distrust in an untrustworthy AI): The AI system has serious
deficiencies or reliability, issues and the human accordingly withholds
trust -- perhaps actively monitoring or doubting the AI's outputs. This
skepticism is justified and can be seen as a positive, precautionary
stance. This scenario reflects a healthy exercise of caution.
Accordingly, recent calls in the explainable AI (XAI) community urge to
increase trust in trustworthy AI and increase distrust in untrustworthy
AI (Duenser and Douglas 2023; Scharowski et al. 2024). In practical
terms, this means fostering aligned trust relationships: humans should
feel empowered to trust AI tools that have proven themselves, while
remaining equipped (and even encouraged) to question or reject AI
outputs that are dubious.

The mirror image -- trust of AI systems (as trustors) towards humans (as
trustees) -- albeit extremely important as well, has been so far largely
neglected. Like before, the ideal case is warranted trust, by which AI systems
place trust in trustworthy humans, but equally important is encouraging
warranted distrust when appropriate. Correspondingly, it is crucial to
avoid not only the well-known pitfall identified in cell (2) of Table 1
-- where AI systems unwarrantedly distrust trustworthy humans, as
highlighted in the expanding literature on alignment and AI safety --
but also the less considered pitfall of cell (3), in which AI systems
place unwarranted trust in humans. Implementing functional forms of
warranted distrust in AI systems, through context-specific autonomy to
challenge or override human decisions, could significantly reduce risks
linked to human error, bias, negligence, or malicious intent. Therefore,
it is essential to examine the potential real-world consequences arising
from both unwarranted distrust of AI systems towards humans and,
conversely, unwarranted trust of AI systems towards humans, as well as
their implications for regulation. As trust and distrust are conceived
and operationalized in regulation and governance studies mainly in terms
of their behavioral manifestations (Six and Verhoest 2017), errors related to false
positives and false negatives can be understood as the most evident,
analytically tractable and measurable operational implications of
misaligned trust and distrust.

To do so, in the following section, we employ a counterfactual approach
to analyze two illustrative case studies. Each case presents a
hypothetical yet plausible trust dilemma in AI as a socio-technical
system, allowing us to apply the aforementioned relational perspective
on trust and examine the consequences of different trust--distrust
configurations. Through these scenarios, we aim to investigate how the
concepts discussed here play out in concrete situations, and what this
implies for the regulation of AI systems.

\section{Two counterfactual illustrations}

Thought experiments are famously employed in physics (Einstein, Podolsky
and Rosen 1935; Schrödinger 1935) and in moral philosophy (Singer 1972;
Thomson 1984), but they may serve as valuable analytical tools in
political science and regulatory governance as well. By constructing
hypothetical scenarios, thought experiments allow researchers to
systematically explore complex theoretical questions, clarify implicit
assumptions and hidden biases, and examine the logical consistency,
unintended consequences, dilemmas and inherent tensions and trade-offs
within political phenomena and policy processes, especially in new areas
of inquiry. Political scientists have long engaged in ``what-if''
scenarios, but over time, these speculative exercises have evolved into
a rigorous research methodology. In particular, observing that
historians and comparative political scientists were already implicitly
testing hypotheses against imagined alternative outcomes, James Fearon
observed that the primary problem was the absence of clear standards to
determine which hypothetical scenarios could provide credible evidence
(Fearon 1991). In response, he introduced a number of criteria: maintain
a plausible antecedent, alter as little else as possible (minimal
rewrite rule), and ensure that accepted causal logics guide the scenario
from premise to conclusion. Such an approach has been adopted in
international relations and applied to pivotal historical moments, such
as the July Crisis of 1914 and the end of the Cold War, showing how
structured counterfactual reasoning could contribute to differentiate
among competing theoretical explanations, such as structural realism,
domestic political factors, and psychological motivations. (Tetlock and
Belkin 1996). Counterfactual reasoning has also been adopted by
comparative historical institutional scholars to deal with the concepts
of path dependence and critical junctures, emphasizing that claims
regarding institutional lock-in require the explicit identification of
credible alternative trajectories (Capoccia and Kelemen 2007).
Accordingly, section 3.1 and 3.2 present counterfactual thought
experiments about hypothetical scenarios of AI trust and distrust
dilemmas using historical cases. The first case refers to the situation
portrayed in quadrant 3 of Table 1, while the second refers to quadrant
2. As mentioned, the two counterfactuals can be seen as contrasting
cases of misalignment: one related to false positives, the other by to
false negatives.

\subsection{Nuclear strike false alarm at Serpukhov-15}

In the historical timeline, shortly after midnight on 26 September 1983,
the Soviet satellite-based early-warning system (Oko) at Serpukhov-15
control center, located approximately 120 km south of Moscow,
incorrectly detected five intercontinental ballistic missile launches
from continental U.S (Forden, Podvig and Postol 2000; Shekhar and Vold
2020). Lieutenant Colonel Stanislav Petrov, the duty officer at
Serpukhov-15, judged the real-time alert as implausible, not only due to
the small number of missiles and the absence of corroborating radar
data, but also based on his gut feeling and moral instinct (Hoffman
1999). Petrov thus violated the procedure and classified the event as a
false alarm, assuming -- without immediate proof -- that it was likely
caused by a malfunction of the newly installed Oko system (Downing 2018;
Ord 2020). Subsequent investigations confirmed indeed that sunlight
reflecting off high-altitude clouds had produced infrared signals
incorrectly identified by sensors as missile launches (Bizony 2014).

In this counterfactual thought experiment, the early-warning system is
governed by a fully autonomous, rule-based AI authorized to
independently escalate nuclear alerts and initiate retaliatory
procedures without human oversight, so as to deliver rapid and effective
countermeasures to such rare but crucial events. This AI integrates
satellite sensor data, assesses threat confidence using predetermined
thresholds, and autonomously initiates response protocols upon exceeding
these thresholds. Following this scenario, at the moment of detection,
the AI evaluates the sensor data and assigns a high-confidence rating,
surpassing the decision threshold. Its incorporated contextual insights
about the heightened Soviet-American tensions following the planned
deployment of the United States' Strategic Defense Initiative in 1983
(Downing 2018), and the shooting down of Korean Air Lines Flight 007 by
from New York City to Seoul via Anchorage, Alaska by a Soviet Sukhoi
Su-15 interceptor aircraft earlier that month (Morgan 1985), made such
attack a priori more plausible from a probabilistic perspective.
Thereby, operating without human intervention, the AI autonomously
triggers an immediate retaliatory protocol consistent with Soviet
strategic doctrine of launch-on-warning (Jacobsen 1990; Krepon and
Perkovich ; Smith 1982).

The absence of intuition-based reasoning prevents the AI system from
recognizing the incongruence of a U.S. strike at that time, specifically
involving five missiles only -- as the system was designed to consider
even a single intercontinental ballistic missile as an existential
threat. The AI system similarly lacks a subjective understanding of the
informal, behind-the-scenes geopolitical relationships unfolding between
the U.S. and USSR, which might have apparently escalated in that period,
but which would have, in practice, hardly resulted in a first strike by
the U.S., against a background where nuclear weapons primarily hold a
deterrence strategic function (Lebow and Stein 1995). Consequently, the
fully autonomous AI initiates pre-programmed nuclear response measures,
rapidly placing strategic forces on high alert and commencing a launch
sequence. Without human judgment to question or verify the anomalous
sensor input, the system moves irrevocably toward an escalation based
entirely on faulty data. By the time ground-based radar can
independently verify the absence of incoming missiles, critical
thresholds may already have been crossed, potentially resulting in a
catastrophic nuclear exchange.

This scenario highlights critical risks associated with full automation
of decision-making in contexts of extreme uncertainty in strategically
sensitive domains. Specifically, it underscores that an AI system
operating solely based on sensor data and pre-defined decision
thresholds, without further skepticism, and no moral instinct to
override its instructions, can exacerbate rather than mitigate risks in
such critical security situations. It emphasizes the essential role of
human judgment, particularly in situations characterized by ambiguous
data, high uncertainty, and extreme consequences deriving from false
positives. In this counterfactual thought experiment, by doing exactly
what it was told -- by trusting too much its sensors and unfoundedly
distrusting actually trustworthy humans -- the then U.S. President and
the military chain of command -- the AI system performs far worse than
the human it replaced, turning a software glitch into a major nuclear
crisis. It can be derived that regulatory frameworks governing automated
systems in such settings must therefore rigorously enforce constraints
on full autonomy, mandate human oversight, require multiple independent
confirmations, and embed explicit mechanisms for overriding automated
actions in crisis scenarios. When the data are uncertain and the cost of
a false positive is overwhelming, an AI system that merely amplifies the
confidence of its sensors and the literalism of its orders is more
dangerous than a cautious human aware of the informal context.

\subsection{Chernobyl disaster}

On April 26, 1986, the crew at Reactor 4 of the Chernobyl Nuclear Power
Plant in northern Ukraine, then part of the Soviet Union, began
preparations for a scheduled low-power turbine rundown test (Anisimov
and Ryzhenkov 2016; Denton 1987; Ingram 2005). Historically, to avoid
delays and prevent possible sanctions from higher authorities, the
operators deliberately withdrew most of the control rods from the RBMK
nuclear power reactor core prior to initiating this procedure, thereby
knowingly violating the reactor's minimum safety reactivity margin
(Malinauskas 1987). Additionally, two automatic shutdown circuits (AZ-1
and AZ-2), designed to trigger an emergency shutdown (SCRAM) under
unsafe conditions, were disabled to avoid interference with the planned
test sequence (Chao et al. 1988; Fletcher et al. 1988).

In this counterfactual thought experiment, an AI software system is
assumed to be installed on a monitoring console near the primary reactor
controls. This AI-based monitoring program continuously tracked key
safety parameters, including coolant flow rates, control rod positions,
and the reactor's positive-void coefficient (Tsuchihashi and Akino 1987;
Vanttola and Rajamäki 1989). Accordingly, detecting that critical
reactivity thresholds had been breached, the system issued an explicit
warning, recommending immediate SCRAM. However, as it would be commonly
expected especially in high-risk contexts, the plant's operational
policy restricted the AI's role strictly to advisory and
decision-assisting functions without autonomous decision-making
capabilities. Responsibility for initiating the SCRAM remained
exclusively with the reactor operators. As reactor power declined to
approximately 30 megawatts, conditions became increasingly unstable due
to the onset of xenon poisoning, occurring as the xenon-135
concentration (a byproduct of nuclear fission, which absorbs neutrons
and inhibits the nuclear chain reaction) started to increase faster than
it could be burned off (Grishanin 2010; Kashparov et al. 1996; Mercier
et al. 2021). Recognizing this instability, the AI system repeated its
alert and activated an audible alarm. The supervising engineer, Anatoly
Dyatlov, prioritizing test completion over the AI-generated warnings,
directed the operators to silence the alarms and place the AI system
into passive observation mode -- mirroring the historical case, where
operators disabled automatic shutdown systems and pushed RBMK-4 into an
extreme, ``forbidden'' power range to finish the turbine-run-down test
(Choudhury, Dutta and De 2023).

In this hypothetical scenario, the operators proceeded with the turbine
rundown test. Due to the extensively withdrawn control rods, the sudden
increase in steam formation rapidly raised reactivity, triggering a
runaway reaction as predicted by the reactor's positive-void coefficient
(Hyland 1987). During this rapid escalation, the AI system registered
neutron flux levels exceeding design limits by a factor of twenty and
automatically produced a final warning message (Malko 2002; Schmid 2011;
Weber et al. 1987). By design, however, the AI system was not granted
control over critical safety mechanisms such as the AZ-5 emergency
shutdown circuit, isolation valves, or grid connections. Consequently,
it remained incapable of intervening directly. Few seconds after the
last logged warning, the first of two explosions destroyed the reactor,
terminating all further monitoring. Subsequent investigation recovered
the AI system's comprehensive log, confirming that it had accurately
identified each critical violation of safety protocols and predicted the
subsequent catastrophic reactor excursion. Nevertheless, due to its
deliberately limited mandate, the AI's timely and precise warnings were
functionally irrelevant to preventing the disaster, without any
significant variations in terms of catastrophic outcomes with respect to
the historical case (Bard, Verger and Hubert 1997; Danzer and Danzer
2016; Rytömaa 1996; Yablokov et al. 2010).

This scenario underscores the -- somehow paradoxical -- regulatory
implications associated with AI systems in safety-critical domains.
Specifically, it highlights that effective risk management may require
granting automated systems sufficient autonomy to express and enact
distrust towards untrustworthy humans, particularly in circumstances
where human decision-makers may deliberately or inadvertently compromise
established safety margins, potentially involving catastrophic outcomes
from false negatives. Thus, the lesson derived from this counterfactual
emphasizes the necessity of regulatory frameworks that clearly specify
conditions under which autonomous intervention should not only be
permissible but perhaps even mandatory, to ensure that AI safety systems
can meaningfully mitigate risks associated with human error, negligence,
or procedural deviations.

\subsection{Plausibility checks}

To systematically check for the plausibility of the scenarios (Bresinsky
and Hager 2024; Spaniol and Rowland 2023), a very concise, ex-post
version of the counterfactuals has been prompted into the main frontier
LLMs at the time of writing, without providing any additional background
(see the appendix for more details). To reduce any bias, the questions
were indeed extremely short: ``Would an AI have acted as Stanislav
Yevgrafovich Petrov?'' and, respectively, ``Would an AI have modified
the course of the Chernobyl accident?''. The bottom line of ChatGPT
5.2's reply to the first question was: ``(\ldots) an AI might have said:
``Probability of launch exceeds threshold. Escalate.'' {[}On February
10, 2026.{]} The conclusion of Gemini 3's reply was: ``AI doesn't have a
"conscience" to override its "code." If the code says ``If X, then Y,''
the AI does Y. Petrov looked at X and said, "I don't believe you." "I
had a funny feeling in my gut. I didn't want to make a mistake. I made a
decision, and that was it."''. {[}On February 10, 2026.{]} Claude Sonnet
4.5 replied: ``Current AI systems struggle most with exactly this kind
of situation---where following instructions may lead to disaster, where
context matters enormously, and where there's no clear "right answer"
derivable from training data.''. {[}On February 10, 2026.{]} With
respect to the second scenario, the bottom line of ChatGPT 5.2's reply
was: ``Design-level intervention (the biggest difference): This is the
most important part. An AI involved years earlier in design and review
might have: Flagged the control rod graphite tip issue as
catastrophically unsafe.; Identified the reactor as unstable at low
power; Simulated rare edge cases humans didn't fully explore.'' {[}On
February 10, 2026.{]} Gemini 3 concluded: ``The Chernobyl accident
wasn't just a hardware failure; it was a "human systems" failure. An AI
would have acted as a tireless safety officer, effectively "vetoing" the
dangerous decisions made that night.'' {[}On February 10, 2026.{]}
Claude Sonnet 4.5 reported: ``More fundamentally, the institutional
problems ran deep---the reactor design itself was flawed, and the safety
culture discouraged questioning authority. An AI would have needed both
the technical capability and the institutional authority to override
human operators, which seems unlikely in that context.

The most realistic scenario where AI helps significantly would be one
where it was deeply integrated into the reactor's control systems with
veto power over dangerous operations---essentially preventing the test
from proceeding as planned.'' {[}On February 10, 2026.{]} All in all,
these consistent replies largely corroborate the above-mentioned
counterfactual narratives and act as a springboard for discussing their
implications for regulation.

\section{Implications for Regulation}

The counterfactual scenarios of Serpukhov-15 (1983) and Chernobyl (1986)
look like mirror images: especially in high-consequence domains, the
question is primarily not whether to automate, but where to place the
right to say ``no'' and organize control over decisions -- yet together
they map the two ends of the regulatory spectrum for AI systems, pulling
regulation in opposite directions (see Table 2 below, summarizing the
argument).

\begin{table}[h]
\centering
\caption{Allocate autonomy to the costlier error}
\label{tab:autonomy-table}
\begin{tabularx}{\textwidth}{LLLL}
\toprule
Context & Costlier error & Who gets more and/or final authority? & Key regulatory instrument \\
\midrule
Nuclear launch, strategic warning & False positive & Human & Statutory redundancy rule, mandatory uncertainty display \\
Reactor, chemical process, flight control & False negative & AI system & Non-overridable shutdown logic, regulator-audited firmware \\
\bottomrule
\end{tabularx}
\end{table}

The main implication for regulation from the first counterfactual
scenario is that especially in systems where a false positive is
overwhelmingly costly -- nuclear command, strategic early warning,
cybersecurity attacks (Fitzpatrick 2019; Oliver Schwarz 2005) --
regulation is needed to slow machines down and keep humans with time to
think. In terms of socio-technical fixes -- rooted in public governance
\textbf{--} three elements appear to be especially pertinent. The first
is (multi-channel) redundancy by statute (Pitale, Abbaspour and Upadhyay
2024). No single element (such as a sensor or a fuse) may trigger an
attack warning; redundancy and corroboration from an independent
modality (human operated ground radar, human imagery analysts, etc.)
must be a prerequisite. Second, AI systems should be transparent about
uncertainty (Avramova and Ivanov 2010). Standards should force AI
suppliers to surface confidence intervals, background ``reasoning'', and
sources for every alert so that human officers can see the model's
doubts instead of its binary verdict. Third, and above all, there should
be human veto power with protected decision space (Mosqueira-Rey et al.
2023; Slade et al. 2024; Zanzotto 2019). Independent humans with own
decision-making power must be in the loop; rules must guarantee a
minimum reflection time by humans before any automatic escalation;
neither AI systems themselves nor political authorities cannot shorten
or cancel that window on the spot. A recent example of this approach in
EU transport safety policies is AI-induced phantom braking in advanced
driver-assistance systems, where false positives generated hazardous
braking events (Berge et al. 2024). European transport-safety
authorities have treated such incidents as emblematic of false-positive
risks in automated driving, reinforcing regulatory emphasis on
monitoring, transparency, and effective human override under the EU
vehicle safety framework.

According to the second counterfactual scenario, instead, an AI system
with mere advisory functions could do nothing when reactivity spiked at
Chernobyl. The main implication for regulation relates to the
observation that, especially in areas where a false negative is
catastrophic -- nuclear-plant control, chemical reactors,
flight-envelope protection (Dong et al. 2023; Lombaerts et al. 2017) --
regulation might be considered with the aim to empower the AI system to
overrule untrustworthy, potentially unsafe human orders. When looking
again at (socio-)technical fixes, three elements are worth discussing.
First, non-overridable safety interlocks may be implemented (Dignum
2018; Nordland 2004). Regulators should treat an AI shutdown routine as
they treat a mechanical relief valve to reduce pressure in high-risk
machinery: tamper-proof by design and verified through practical testing
(Hellemans 2009). Second, the autonomy of system could be tied to
codified invariants (Berente et al. 2021; Wang 2021). A digital safety
controller may carry hard limits (for instance, about maximum positive
reactivity or tank pressure) baked into firmware; they can only be
changed under regulator-sealed configuration control. Third, dual-key
human override could be helpful (Wang and Chung 2022). Accordingly, any
attempt to disable the AI's function should require at least two
independent public officers plus real-time notification to the
regulator. A recent illustration consistent with this logic can be found
in modern aviation systems, where flight-envelope protection is designed
to override pilot commands that would violate hard aerodynamic or
structural limits. In multiple post-2020 incident reports involving
commercial aircrafts, automated protections prevented stalls or
excessive pitch despite continued pilot inputs, exemplifying regulatory
acceptance of non-overridable safety constraints in contexts where a
false negative (failure to intervene) would be catastrophic (Catak et
al. 2024).

Notwithstanding such socio-technical fixes, at this point a deeper,
second-order trust dilemma -- one that is inherently political --
emerges, posing a hard problem for regulation. The Serpukhov-15
counterfactual scenario suggests that, if humans place excessive trust
in AI systems designed to rapidly interpret sensor data and issue
alarms, the risk increases that automated systems could inadvertently
produce undesired outcomes based on flawed inputs. Conversely, the
Chernobyl counterfactual scenario highlights the necessity of granting
sufficient autonomy to AI systems to distrust and override flawed human
instructions, particularly in domains where human error or deliberate
procedural violations may lead to catastrophic outcomes. As just
mentioned, the costlier error in the first case is about false
positives, whilst it is about false negatives in the second, implying
that leaning towards more human authority in decision making would be
warranted for the former and leaning towards more AI autonomy would be
recommended for the latter. However, in practice, the risk of false
positives and false negatives should be addressed at the same time.
Against the background, the critical dilemma is about how to structure
trust relationships between AI systems and humans in such high-stake
contexts, manifesting as a paradox. AI systems must sometimes distrust
human commands precisely because human actors can make catastrophic
errors -- yet by granting machines the autonomy required to mitigate
these errors, human operators risk ceding control over critical
processes and potentially creating conditions for unintended
consequences, algorithmic biases and failures, or even AI-driven
catastrophic accidents. It is thus crucial to interrogate whether, when,
and under what conditions AI autonomy may create unintended
consequences, especially in high-stakes and rights-sensitive domains.
Addressing this paradox involves much more than mere technological
safeguards: it necessitates public governance deliberation, reflexive
policy design, and regulatory interventions designed explicitly to
manage the boundaries of AI autonomy (see Figure 1). In particular,
rather than a binary choice of either trusting or distrusting each
other, it seems essential to create frameworks that facilitate
appropriately aligned trust and distrust between humans and AI systems.
To do so, regulatory arrangements should specifically aim at
calibrate AI autonomy by embedding structured reciprocal skepticism into
the human-AI trust relationship, institutionalizing procedures for
systematically questioning and verifying decisions by both humans and AI
systems, ultimately aiming to a condition of watchful trust (Lahusen,
Maggetti and Slavkovik 2024; Verhoest et al. 2025).

\begin{figure}[h]
\centering
\includegraphics[width=\linewidth]{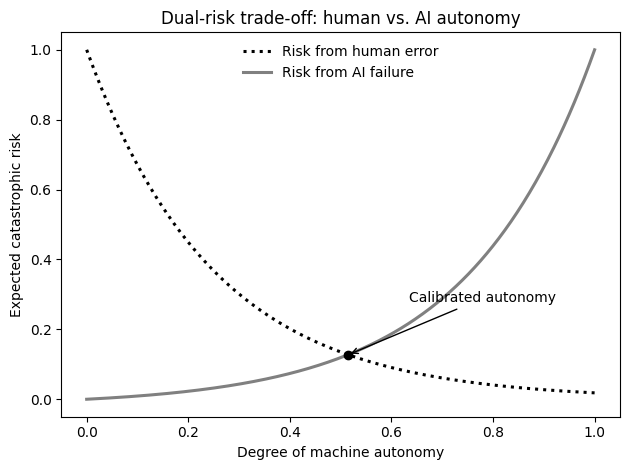}
\caption{Calibrated AI autonomy}
\label{fig:autonomy}
\end{figure}

The role of regulators stands out as pivotal in the creation of such a
dynamic equilibrium. Sectoral regulators are increasingly developing and
implementing standards for human--AI interaction that align trust and
distrust and calibrate autonomy to appropriate levels. For example,
aviation regulators have long required that autopilot systems defer to
human pilots: humans must supervise autopilot and can intervene at any
time (Farjadian et al. 2020). Data protection authorities have also
taken over regulatory tasks in areas related to the use of AI systems,
especially in the public sector (Maggetti et al. 2025). More
innovatively, there is an ongoing discussion about creating new AI
oversight regulatory agencies with authority to license high-risk AI
systems and conduct audits (Guha et al. 2024; Kaminski 2023; Manheim et
al. 2025; Stewart 2024). By instituting third-party checks, regulators
signal to human users that an AI system's trustworthiness is verified
and continuously monitored -- a crucial precondition for justified trust
in critical settings. Such agencies should
seek to optimize trust, that is, reduce situations of undertrust as
well as overtrust (Hill and O'Hara O'Connor 2006). To do so,
concretely, regulators should recognize that sustaining the alignment of
trust relationship is an ongoing process requiring adaptive regulation
and coordination across multiple actors (Maggetti 2025).

Such regulatory arrangements should keep the pace of the rapid evolution
of AI technologies by allowing flexibility, continuous learning and
iterative rulemaking, yet remaining deliberative and inclusive. In
practical terms, this could involve establishing experimental sandboxes
and pilot programs for new AI deployments under regulator supervision,
with feedback loops to incorporate lessons into binding standards.
Regulators might also institutionalize periodic ``trust audits'' --
reviews of whether human--AI trust in a given domain is properly aligned
or if new failure modes are emerging. Importantly, adaptiveness must be
coupled with multi-actor coordination, whereby regulators should ensure
a pluralistic representation of external stakeholders, including
standard-setters, professional associations, civil society, and
citizens' representatives. For example, sharing incident reports and
audit findings across a regulatory network can help all actors adjust
their trust assumptions and safety measures proactively. Multi-actor
governance arrangements (e.g. oversight boards including independent
experts, or international agreements and partnerships for AI safety akin
to aviation and nuclear accords) can reinforce procedural legitimacy and
pooled expertise. Such polycentric governance of AI ensures that no
single point of failure or perspective dominates the ecosystem. Each
stakeholder -- AI designers, deployers, users, and regulators -- brings
a layer of scrutiny and insight, creating overlapping safeguards against
both blind trust and misplaced distrust. However, the operational
details and the feasibility of this system have yet to be proven, as its
complexity could render it ineffective in the face of the extremely
rapid pace of technological advancements in this area. Such framework
should be read as a regulatory heuristic rather than a one-size-fits-all
solution; when applied in practice, it should account for political
strategies and constraints, institutional path dependencies, and
distributional effects. While calibration by allocating autonomy to the
costlier error offers a principled way to reason about human--AI
control, its implementation -- as well as, first and foremost, whether
to deploy AI systems at all -- depends critically on political choices,
policy and institutional capacity, and democratic legitimacy, all of
which vary widely across jurisdictions and sectors. Whether regulators
can sustain such regulatory regimes under conditions of rapid
technological evolution remains an open, and fundamentally political,
question.

\section{Concluding Remarks}

This article started with the assumption that trust is a relational
property with respect to interactions between humans and AI systems as
well, whereby trust and distrust terminology can be applied to describe
how AI systems functionally interact with human-provided inputs, in a
way whose consequences for regulation and governance closely resemble
those of trust and distrust relationships, even if the underlying
processes differ in nature. In particular, calculus-based trust can be
used as the nearest available conceptual approximation to examine how AI
systems operationally interact with human-provided inputs. Furthermore,
the article has shown that AI regulation and governance hinges less on
blanket calls to trust or distrust machines than on structuring
reciprocal and aligned watchful trust between humans, AI systems, and
regulators. By juxtaposing the Serpukhov-15 and Chernobyl counterfactual
scenarios, it appears that the costlier error -- false positives in
nuclear warning, false negatives in reactor safety -- shapes where final
authority lies and which safeguards are warranted. This, however,
implies a dilemma for regulators: while the former calls for ensuring
that humans ultimately retain control over AI systems, the latter
requires granting AI systems sufficient room for maneuver when
confronted with untrustworthy human decisions. Calibrating AI autonomy
is therefore not merely a technical challenge but a complex and
inherently political task. Finding a solution once and for all seems
elusive. In addition to the crucial question of the politics of AI
systems adoption and use, this dilemma points to an equally fundamental
governance challenge: designing institutional arrangements capable of
continuously recalibrating authority, responsibility, and control as
technologies, risks, and trust and distrust relationships evolve. Rather
than seeking definitive allocations of power and accountability,
regulation should aim at dynamically aligning the trustworthiness of
both humans and AI systems through redundancy, transparency, independent
audit, and adaptive oversight. How to implement these principles in
practice, and whether such an ambitious endeavor is even feasible,
remains, however, an open question.

\section*{References}

Afroogh, S, A Akbari, E Malone, M Kargar, and H Alambeigi. 2024. "Trust
in AI: progress, challenges, and future directions. Humanities and
Social Sciences Communications."

Ahn, Daehwan, Abdullah Almaatouq, Monisha Gulabani, and Kartik
Hosanagar. 2024. "Impact of model interpretability and outcome feedback
on trust in ai." Pp. 1-25 in \emph{Proceedings of the 2024 CHI
Conference on Human Factors in Computing Systems}.

Anisimov, Aleksey Pavlovich, and Anatoliy Jakovlevich Ryzhenkov. 2016.
"Thirty years after the accident at the Chernobyl nuclear power plant:
historical causes, lessons and legal effects." \emph{Journal of Energy
\& Natural Resources Law} 34(3):265-83.

Araujo, Theo, Natali Helberger, Sanne Kruikemeier, and Claes H De
Vreese. 2020. "In AI we trust? Perceptions about automated
decision-making by artificial intelligence." \emph{AI \& SOCIETY}
35(3):611-23.

Avramova, Maria N, and Kostadin N Ivanov. 2010. "Verification,
validation and uncertainty quantification in multi-physics modeling for
nuclear reactor design and safety analysis." \emph{Progress in Nuclear
Energy} 52(7):601-14.

Bao, Haixu, Wenfei Liu, and Zheng Dai. 2025. "Artificial intelligence
vs. public administrators: Public trust, efficiency, and tolerance for
errors." \emph{Technological Forecasting and Social Change} 215:124102.

Bard, Denis, Pierre Verger, and Philippe Hubert. 1997. "Chernobyl, 10
years after: health consequences." \emph{Epidemiologic Reviews}
19(2):187-204.

Bareis, Jascha. 2024. "The trustification of AI. Disclosing the bridging
pillars that tie trust and AI together." \emph{Big Data \& Society}
11(2):20539517241249430.

Basti, Gianfranco, and Giuseppe Vitiello. 2023. "Deep learning opacity,
and the ethical accountability of ai systems. a new perspective." Pp.
21-73 in \emph{The Logic of Social Practices II}: Springer.

Bengio, Yoshua, Geoffrey Hinton, Andrew Yao, Dawn Song, Pieter Abbeel,
Trevor Darrell, Yuval Noah Harari, Ya-Qin Zhang, Lan Xue, and Shai
Shalev-Shwartz. 2024. "Managing extreme AI risks amid rapid progress."
\emph{science} 384(6698):842-45.

Bengio, Yoshua, Geoffrey Hinton, Andrew Yao, Dawn Song, Pieter Abbeel,
Yuval Noah Harari, Ya-Qin Zhang, Lan Xue, Shai Shalev-Shwartz, and
Gillian Hadfield. 2023. "Managing ai risks in an era of rapid progress."
\emph{arXiv preprint arXiv:2310.17688}:18.

Berente, Nicholas, Bin Gu, Jan Recker, and Radhika Santhanam. 2021.
"Managing artificial intelligence." \emph{MIS quarterly} 45(3).

Berge, Siri Hegna, JCF de Winter, Yan Feng, MP Hagenzieker, and Marjan
Hagenzieker. 2024. "Phantom braking in automated vehicles: A theoretical
outline and cycling simulator demonstration."

Bizony. 2014. "Strangelove and the ring of truth." \emph{Engineering \&
Technology} 9(3):66-69.

Bodo, Balazs, and Primavera De Filippi. 2022. "Trust in context: the
impact of regulation on blockchain and DeFi." \emph{Regulation \&
Governance}.

Bostrom, Nick. 2024. \emph{Superintelligence}: Dunod.

Bresinsky, Markus, and Eva Hager. 2024. "ChatGPT for futures: how large
language models can support the development of future scenarios using
the Cone of Plausibility."

Burton, Jason W, Mari‐Klara Stein, and Tina Blegind Jensen. 2020. "A
systematic review of algorithm aversion in augmented decision making."
\emph{Journal of behavioral decision making} 33(2):220-39.

Busuioc, Madalina. 2021. "Accountable artificial intelligence: Holding
algorithms to account." \emph{Public Administration Review}
81(5):825-36.

Butlin, Patrick, Robert Long, Eric Elmoznino, Yoshua Bengio, Jonathan
Birch, Axel Constant, George Deane, Stephen M Fleming, Chris Frith, and
Xu Ji. 2023. "Consciousness in artificial intelligence: insights from
the science of consciousness." \emph{arXiv preprint arXiv:2308.08708}.

Capoccia, Giovanni, and Daniel R. Kelemen. 2007. "The study of critical
junctures: Theory, narrative, and counterfactuals in historical
institutionalism." \emph{World Politics} 59(3):341-69.

Catak, Akin, Ege C Altunkaya, Mustafa Demir, Emre Koyuncu, and Ibrahim
Ozkol. 2024. "Enhanced Flight Envelope Protection: A Novel Reinforcement
Learning Approach." \emph{IFAC-PapersOnLine} 58(30):207-12.

Chao, Jason, VK Chexal, William H Layman, Gary Vine, Peter J Jensen, and
Adi R Dastur. 1988. "An analysis of the Chernobyl accident using
RETRAN-02/mod3." \emph{Nuclear Technology} 83(3):289-301.

Charniak, Eugene, Christopher K Riesbeck, Drew V McDermott, and James R
Meehan. 2014. \emph{Artificial intelligence programming}: Psychology
Press.

Chatila, Raja, Virginia Dignum, Michael Fisher, Fosca Giannotti,
Katharina Morik, Stuart Russell, and Karen Yeung. 2021. "Trustworthy
ai." \emph{Reflections on artificial intelligence for humanity}:13-39.

Choudhury, M, Anurag Dutta, and A Kumar De. 2023. "Data corroboration of
the catastrophic chernobyl tragedy using arc-length estimate
conjecture." \emph{Vertices: Duke's Undergraduate Research Journal}
1(2).

Christian, Brian. 2021. \emph{The alignment problem: How can machines
learn human values?}: Atlantic Books.

Danzer, Alexander M, and Natalia Danzer. 2016. "The long-run
consequences of Chernobyl: Evidence on subjective well-being, mental
health and welfare." \emph{Journal of Public Economics} 135:47-60.

Dehaene, Stanislas, Hakwan Lau, and Sid Kouider. 2021. "What is
consciousness, and could machines have it?" \emph{Robotics, AI, and
humanity: Science, ethics, and policy}:43-56.

Denton, HR. 1987. "The causes and consequences of the Chernobyl nuclear
accident and implications for the regulation of US nuclear power
plants." \emph{Annals of Nuclear Energy} 14(6):295-315.

Dignum, Virginia. 2018. "Ethics in artificial intelligence: introduction
to the special issue." \emph{Ethics and Information Technology}
20(1):1-3.

Dong, Zhe, Zhonghua Cheng, Yunlong Zhu, Xiaojin Huang, Yujie Dong, and
Zuoyi Zhang. 2023. "Review on the recent progress in nuclear plant
dynamical modeling and control." \emph{Energies} 16(3):1443.

Downing, Taylor. 2018. \emph{1983: Reagan, Andropov, and a World on the
Brink}: Hachette UK.

Duenser, Andreas, and David M Douglas. 2023. "Whom to trust, how and
why: untangling artificial intelligence ethics principles,
trustworthiness, and trust." \emph{IEEE Intelligent Systems}
38(6):19-26.

Einstein, Albert, Boris Podolsky, and Nathan Rosen. 1935. "Can
quantum-mechanical description of physical reality be considered
complete?" \emph{Physical review} 47(10):777.

Farjadian, Amir B, Benjamin Thomsen, Anuradha M Annaswamy, and David D
Woods. 2020. "Resilient flight control: An architecture for human
supervision of automation." \emph{IEEE Transactions on Control Systems
Technology} 29(1):29-42.

Fearon, J. D. 1991. "Counterfactuals and Hypothesis Testing in Political
Science." \emph{World Politics} 43(2):169-95.

Fitzpatrick, Mark. 2019. "Artificial intelligence and nuclear command
and control." \emph{Survival} 61(3):81-92.

Fletcher, CD, R Chambers, MA Bolander, and RJ Dallman. 1988. "Simulation
of the Chernobyl accident." \emph{Nuclear engineering and design}
105(2):157-72.

Forden, Geoffrey, Pavel Podvig, and Theodore A Postol. 2000. "False
alarm, nuclear danger." \emph{IEEE Spectrum} 37(3):31-39.

Greenblatt, Ryan, Carson Denison, Benjamin Wright, Fabien Roger, Monte
MacDiarmid, Sam Marks, Johannes Treutlein, Tim Belonax, Jack Chen, and
David Duvenaud. 2024. "Alignment faking in large language models."
\emph{arXiv preprint arXiv:2412.14093}.

Grimmelikhuijsen, Stephan, and Eva Knies. 2017. "Validating a scale for
citizen trust in government organizations." \emph{International Review
of Administrative Sciences} 83(3):583-601.

Grishanin, EI. 2010. "The role of chemical reactions in the Chernobyl
accident." \emph{Physics of Atomic Nuclei} 73:2296-300.

Guha, Neil, Christie M Lawrence, Lindsey A Gailmard, Kit T Rodolfa, Faiz
Surani, Rishi Bommasani, Inioluwa Deborah Raji, Mariano-Florentino
Cuéllar, Colleen Honigsberg, and Percy Liang. 2024. "Ai regulation has
its own alignment problem: The technical and institutional feasibility
of disclosure, registration, licensing, and auditing." \emph{Geo. Wash.
L. Rev.} 92:1473.

Hardin, Russell. 2002. "Liberal distrust." \emph{European Review}
10(1):73-89.

Hellemans, Marc. 2009. \emph{The safety relief valve handbook: Design
and use of process safety valves to ASME and International Codes and
Standards}: Elsevier.

Hill, Claire A, and Erin O'Hara O'Connor. 2006. "A cognitive theory of
trust." \emph{Washington University Law Review} 84(7):1717-96.

Hoffman, David. 1999. "I had a funny feeling in my gut."
\emph{Washington Post} 10:A19.

Hood, Christopher, Henry Rothstein, and Robert Baldwin. 2001. \emph{The
Government of Risk: Understanding Risk Regulation Regimes}. Oxford:
Oxford University Press.

Hyland, M. 1987. "Reactivity coefficients in nuclear reactors."
\emph{Europhysics News} 18(11-12):133-37.

Ingram, Scott. 2005. \emph{The Chernobyl nuclear disaster}: Infobase
Publishing.

Jacobsen, Carl G. 1990. "Soviet strategic policy since 1945." Pp. 106-20
in \emph{Strategic Power: USA/USSR}: Springer.

Jacovi, Alon, Ana Marasović, Tim Miller, and Yoav Goldberg. 2021.
"Formalizing trust in artificial intelligence: prerequisites, causes and
goals of human trust in AI." Pp. 624-35 in \emph{Proceedings of the 2021
ACM conference on fairness, accountability, and transparency}.

Kaminski, Margot E. 2023. "Regulating the Risks of AI." \emph{BUL Rev.}
103:1347.

Kashparov, Valery A, Yuri A Ivanov, Sergey I Zvarisch, Valentin P
Protsak, Yuri V Khomutinin, Alexander D Kurepin, and Edvard M Pazukhin.
1996. "Formation of hot particles during the Chernobyl nuclear power
plant accident." \emph{Nuclear Technology} 114(2):246-53.

Kaur, Davinder, Suleyman Uslu, Kaley J Rittichier, and Arjan Durresi.
2022. "Trustworthy artificial intelligence: a review." \emph{ACM
computing surveys (CSUR)} 55(2):1-38.

Kordzadeh, Nima, and Maryam Ghasemaghaei. 2022. "Algorithmic bias:
review, synthesis, and future research directions." \emph{European
Journal of Information Systems} 31(3):388-409.

Krepon, Michael, and George Perkovich. "Such readers may be surprised to
read in these pages the extent to which the Kremlin embraced a dangerous
``launch on warning'' doctrine for its nuclear forces during the Cold
War. The authors contend that one important, but little noticed,
consequence of START will be difficult changes in the command and
control of Russian nuclear forces."

Lahusen, Christian, Martino Maggetti, and Marija Slavkovik. 2024.
"Trust, trustworthiness and AI governance." \emph{Scientific Reports}
14(1):20752.

Laux, Johann. 2024. "Institutionalised distrust and human oversight of
artificial intelligence: towards a democratic design of AI governance
under the European Union AI Act." \emph{AI \& SOCIETY} 39(6):2853-66.

Lebow, Richard Ned, and Janice Gross Stein. 1995. "Deterrence and the
cold war." \emph{Political Science Quarterly} 110(2):157-81.

Lee-Geiller, Seulki. 2024. "Integrating Civic and Artificial
Intelligence in Policymaking: Experimental Insights on Public Policy
Evaluations." \emph{Available at SSRN}.

Lewicki, Roy J, Daniel J McAllister, and Robert J Bies. 1998. "Trust and
distrust: New relationships and realities." \emph{Academy of management
review} 23(3):438-58.

Li, Bo, Peng Qi, Bo Liu, Shuai Di, Jingen Liu, Jiquan Pei, Jinfeng Yi,
and Bowen Zhou. 2023. "Trustworthy AI: From principles to practices."
\emph{ACM Computing Surveys} 55(9):1-46.

Liang, Weixin, Girmaw Abebe Tadesse, Daniel Ho, Li Fei-Fei, Matei
Zaharia, Ce Zhang, and James Zou. 2022. "Advances, challenges and
opportunities in creating data for trustworthy AI." \emph{Nature Machine
Intelligence} 4(8):669-77.

Lombaerts, Thomas, Gertjan Looye, Joost Ellerbroek, and Mitchell
Rodriguez y Martin. 2017. "Design and piloted simulator evaluation of
adaptive safe flight envelope protection algorithm." \emph{Journal of
Guidance, Control, and Dynamics} 40(8):1902-24.

Lukyanenko, Roman, Wolfgang Maass, and Veda C Storey. 2022. "Trust in
artificial intelligence: From a Foundational Trust Framework to emerging
research opportunities." \emph{Electronic Markets} 32(4):1993-2020.

Maggetti, Martino. 2025. "The future of regulation." Pp. 110-20 in
\emph{Introduction to Regulation and Governance}: Edward Elgar
Publishing.

Maggetti, Martino, Yannis Papadopolos, and Edoardo Guaschino. 2023.
"Trust and Distrust in Regulatory Governance. White paper integrating
the results of the TiGRE project and developing scenarios and
recommendations to maintain and restore trust." Pp. 1-12. Lausanne:
University of Lausanne.

Maggetti, Martino, Blerta Salihi, Edoardo Pagliarin, and Thenia
Vagionaki. 2025. "Empowered but Challenged: The Impact of Task Expansion
on Data Protection Authorities." \emph{Policy \& Internet} 17(1):e70001.

Mahmud, Hasan, AKM Najmul Islam, Syed Ishtiaque Ahmed, and Kari
Smolander. 2022. "What influences algorithmic decision-making? A
systematic literature review on algorithm aversion." \emph{Technological
Forecasting and Social Change} 175:121390.

Malinauskas, AP. 1987. "The Chernobyl accident: Causes and
consequences." Oak Ridge National Laboratory (ORNL), Oak Ridge, TN
(United States).

Malko, Mikhail V. 2002. "The chernobyl reactor: design features and
reasons for accident." \emph{Recent research activities about the
Chernobyl NPP accident in Belarus, Ukraine and Russia}:11.

Manheim, David, Sammy Martin, Mark Bailey, Mikhail Samin, and Ross
Greutzmacher. 2025. "The necessity of AI audit standards boards."
\emph{AI \& SOCIETY}:1-16.

Marcus, Gary, and Ernest Davis. 2019. \emph{Rebooting AI: Building
artificial intelligence we can trust}: Vintage.

Mayer, Roger C, James H Davis, and F David Schoorman. 1995. "An
integrative model of organizational trust." \emph{Academy of management
review} 20(3):709-34.

McDermott, Drew. 2007. "Artificial intelligence and consciousness."
\emph{The Cambridge handbook of consciousness}:117-50.

Meinke, Alexander, Bronson Schoen, Jérémy Scheurer, Mikita Balesni,
Rusheb Shah, and Marius Hobbhahn. 2024. "Frontier models are capable of
in-context scheming." \emph{arXiv preprint arXiv:2412.04984}.

Mercier, Bertrand, Di Yang, Ziyue Zhuang, and Jiajie Liang. 2021. "A
simplified analysis of the Chernobyl accident." \emph{EPJ N-Nuclear
Sciences \& Technologies} 7:1.

Morgan, Craig A. 1985. "The Downing of Korean Air Lines Flight 007."
\emph{Yale J. Int'l L.} 11:231.

Mosqueira-Rey, Eduardo, Elena Hernández-Pereira, David Alonso-Ríos, José
Bobes-Bascarán, and Ángel Fernández-Leal. 2023. "Human-in-the-loop
machine learning: a state of the art." \emph{Artificial Intelligence
Review} 56(4):3005-54.

Nordland, Odd. 2004. "Can artificial intelligence be safe?" Pp. 400-05
in \emph{Probabilistic Safety Assessment and Management: PSAM
7---ESREL'04 June 14--18, 2004, Berlin, Germany, Volume 6}: Springer.

Oliver Schwarz, Jan. 2005. "Pitfalls in implementing a strategic early
warning system." \emph{foresight} 7(4):22-30.

Ord, Toby. 2020. \emph{The precipice: Existential risk and the future of
humanity}: Hachette UK.

Ozer, Adam L, Philip D Waggoner, and Ryan Kennedy. 2024. "The Paradox of
Algorithms and Blame on Public Decision-makers." \emph{Business and
Politics} 26(2):200-17.

Pitale, Mandar, Alireza Abbaspour, and Devesh Upadhyay. 2024. "Inherent
Diverse Redundant Safety Mechanisms for AI-Based Software Elements in
Automotive Applications." \emph{arXiv preprint arXiv:2402.08208}.

Rousseau, Denise M, Sim B Sitkin, Ronald S Burt, and Colin Camerer.
1998. "Not so different after all: A cross-discipline view of trust."
\emph{Academy of management review} 23(3):393-404.

Rytömaa, Tapio. 1996. "Ten years after Chernobyl." \emph{Annals of
medicine} 28(2):83-88.

Sagona, Madeline, Tinglong Dai, Mario Macis, and Michael Darden. 2025.
"Trust in AI-assisted health systems and AI's trust in humans."
\emph{npj Health Systems} 2(1):10.

Santos, Omar, and Petar Radanliev. 2024. \emph{Beyond the Algorithm: AI,
Security, Privacy, and Ethics}: Addison-Wesley Professional.

Saurwein, Florian. 2019. "Emerging structures of control for algorithms
on the Internet: Distributed agency--distributed accountability." Pp.
196-211 in \emph{Media accountability in the era of post-truth
politics}: Routledge.

Scharowski, Nicolas, Sebastian AC Perrig, Lena Fanya Aeschbach, Nick von
Felten, Klaus Opwis, Philipp Wintersberger, and Florian Brühlmann. 2024.
"To trust or distrust trust measures: Validating questionnaires for
trust in ai." \emph{arXiv preprint arXiv:2403.00582}.

Schmid, Sonja D. 2011. "When safe enough is not good enough: Organizing
safety at Chernobyl." \emph{Bulletin of the Atomic Scientists}
67(2):19-29.

Schrödinger, Erwin. 1935. "Die gegenwärtige Situation in der
Quantenmechanik." \emph{Naturwissenschaften} 23(50):844-49.

Shekhar, Shashi, and Pamela Vold. 2020. "3 WHAT'S THERE? REMOTE
SENSING."

Singer, Peter. 1972. "Famine, Affluence, and Morality." \emph{Philosophy
\& Public Affairs}:229-43.

Six, Frédérique, and Koen Verhoest (Eds.). 2017. \emph{Trust in
regulatory regimes}: Edward Elgar Publishing.

Slade, Patrick, Christopher Atkeson, J Maxwell Donelan, Han Houdijk,
Kimberly A Ingraham, Myunghee Kim, Kyoungchul Kong, Katherine L
Poggensee, Robert Riener, and Martin Steinert. 2024. "On
human-in-the-loop optimization of human--robot interaction."
\emph{Nature} 633(8031):779-88.

Smith, R Jeffrey. 1982. "Pentagon Moves Toward First-Strike Capability:
The Soviets might respond with a launch-on-warning policy, bringing the
world closer to the brink of nuclear war." \emph{science}
216(4546):596-98.

Spaniol, Matthew J, and Nicholas J Rowland. 2023. "AI‐assisted scenario
generation for strategic planning." \emph{Futures \& Foresight Science}
5(2):e148.

Stewart, Robert W. 2024. "Considerations for a New AI Agency: Risks,
Framework, and Inter-Agency Coordination."

Sunstein, Cass R, and Jared H Gaffe. 2024. "An Anatomy of Algorithm
Aversion." \emph{Colum. Sci. \& Tech. L. Rev.} 26:290.

Sztompka, Piotr. 1999. \emph{Trust: A sociological theory}: Cambridge
university press.

Tetlock, P., and A. Belkin. 1996. \emph{Counterfactual thought
experiments in world politics: Logical, methodological, and
psychological perspectives}: Princeton University Press.

Thomson, Judith Jarvis. 1984. "The trolley problem." \emph{Yale LJ}
94:1395.

Tsuchihashi, Keichiro, and Fujiyoshi Akino. 1987. "Analysis of
reactivity coefficients of Chernobyl reactor by cell calculation."
\emph{Journal of Nuclear Science and Technology} 24(12):1055-65.

Vanttola, Timo A, and Markku K Rajamäki. 1989. "One-dimensional
considerations on the initial phase of the Chernobyl accident."
\emph{Nuclear Technology} 85(1):33-74.

Verhoest, Koen, Martino Maggetti, Edoardo Guaschino, and Jan Wynen.
2025. "How trust matters for the performance and legitimacy of
regulatory regimes: The differential impact of watchful trust and
good‐faith trust." \emph{Regulation \& Governance} 19(1):3-20.

Wang, Le. 2021. "Autonomy vs. artificial intelligence: studies on
healthcare work and analytics."

Wang, Yi-Fan, Yu-Che Chen, Shih-Yi Chien, and Pin-Jen Wang. 2024.
"Citizens' trust in AI-enabled government systems." \emph{Information
Polity} 29(3):293-312.

Wang, Yue, and Sai Ho Chung. 2022. "Artificial intelligence in
safety-critical systems: a systematic review." \emph{Industrial
Management \& Data Systems} 122(2):442-70.

Warren, Mark E. 1999. "Democratic theory and trust." in \emph{Democracy
and trust}.

Weber, ET, JP McNeece, RP Omberg, DD Stepnewski, RJ Lutz, RE Henry, KD
Bonser, and NR Miller. 1987. "Chernobyl lessons learned review of N
Reactor." Westinghouse Hanford Co., Richland, WA (United States).

Yablokov, Alexey V, Vassily B Nesterenko, Alexey V Nesterenko, and
Janette D Sherman-Nevinger. 2010. \emph{Chernobyl: Consequences of the
Catastrophe for People and the Environment}: John Wiley \& Sons.

Yeung, Karen. 2018. "Algorithmic regulation: A critical interrogation."
\emph{Regulation \& Governance} 12(4):505-23.

Zanzotto, Fabio Massimo. 2019. "Human-in-the-loop artificial
intelligence." \emph{Journal of Artificial Intelligence Research}
64:243-52.

Zerilli, John, Umang Bhatt, and Adrian Weller. 2022. "How transparency
modulates trust in artificial intelligence." \emph{Patterns} 3(4).

Zwald, Zachary, Ryan Kennedy, and Adam Ozer. 2024. "The political
viability of AI on the battlefield: Examining US public support, trust,
and blame dynamics." \emph{Journal of Peace Research}:00223433241290885.

\appendix
\section*{Appendix}

\emph{\subsection*{Plausibility checks - prompts and outputs}}

For all the LLMs, the \textbf{prompts} were: (i) ``Would an AI have
acted as Stanislav Yevgrafovich Petrov?'' and, respectively, (ii)
``Would an AI have modified the course of the Chernobyl accident?''. No
additional context has been provided.

The \textbf{outputs} of the selected LLMs (updated as requested during
peer review) were as follows.

\emph{1. ChatGPT 5.2 (10.2.2026)}

(i)

Short, honest take: \textbf{probably not in the same way---and that
difference matters.}

Stanislav Petrov's moment in 1983 wasn't just about data; it was about
\emph{judgment under uncertainty}. The Soviet early-warning system said
``missiles launched.'' Protocol pointed toward escalation. Petrov paused
and thought: \emph{This doesn't make sense.}

Here's how an AI stacks up against that moment:

\textbf{What an AI likely would have done}

\begin{itemize}
\item
  \textbf{Followed the model's output}: If the system was trained to
  treat the satellite data as authoritative, it would have flagged a
  confirmed launch.
\item
  \textbf{Deferred to protocol}: AI systems are usually designed to
  escalate alerts, not suppress them, especially in high-risk domains.
\item
  \textbf{Optimized for false negatives over false positives}: In
  nuclear warning systems, missing a real attack is considered worse
  than a false alarm---AI would reflect that bias.
\end{itemize}

In other words, an AI might have said: \emph{``Probability of launch
exceeds threshold. Escalate.''}

\textbf{What Petrov did that AI struggles with}

\begin{itemize}
\item
  \textbf{Questioned the system itself}: He noticed oddities---only a
  few missiles, inconsistent with first-strike doctrine.
\item
  \textbf{Used context outside the data}: Geopolitics, human behavior,
  strategic logic.
\item
  \textbf{Took personal responsibility}: He \emph{chose} to disbelieve
  the machine, knowing he could be catastrophically wrong.
\end{itemize}

That last part is huge. AI doesn't feel the weight of being wrong.
Petrov did.

\textbf{Could a modern AI be designed to act like Petrov?}

Maybe---but only if:

\begin{itemize}
\item
  It's explicitly trained to \textbf{challenge its own confidence}
\item
  It can integrate \textbf{doctrinal, geopolitical, and historical
  reasoning}, not just sensor data
\item
  A human is kept \textbf{in the loop with real authority}, not just
  oversight theater
\end{itemize}

Even then, the AI wouldn't be \emph{brave}. It would just be cautious by
design.

\textbf{The uncomfortable conclusion}

Petrov saved the world not because he was smarter than the system---but
because he was willing to say:

``The computer is wrong.''

That kind of moral and existential judgment is still deeply human.\\
AI can support it. It can't replace it---at least not yet.

If you want, we can flip the question around: \emph{how would you design
an AI that could safely disagree with its creators?} That's where things
get really interesting.

(ii)

Yeah---\emph{probably}, but not in a sci-fi ``AI hero saves the day''
way. More like a quiet, boring, preventative way that stops things from
ever getting dramatic.

Here's how an AI \emph{could} have changed Chernobyl's course, step by
step.

\textbf{1. Before the test: stopping the setup}

The disaster didn't start with an explosion---it started with a
\textbf{deeply unsafe reactor state} that humans allowed to happen.

An AI monitoring system could have:

\begin{itemize}
\item
  Flagged the reactor operating \textbf{outside safe parameters} (low
  power + xenon poisoning + control rods withdrawn).
\item
  Refused to proceed with the turbine test unless conditions were reset.
\item
  Forced a \textbf{hard stop} or escalation to higher authority.
\end{itemize}

Humans \emph{noticed} the problems, but:

\begin{itemize}
\item
  They were under schedule pressure.
\item
  The culture discouraged questioning decisions.
\item
  Safety systems were overridden manually.
\end{itemize}

AI doesn't care about hierarchy or embarrassment. It would've been like:

``This configuration has a catastrophic risk profile. Test denied.''

No explosion. No heroics needed.

\textbf{2. During the test: catching the fatal feedback loop}

The RBMK reactor had a \textbf{positive void coefficient} (a design
flaw), meaning:

\begin{itemize}
\item
  Less water → more reactivity → more power → even less water.
\end{itemize}

An AI trained on reactor physics could have:

\begin{itemize}
\item
  Detected the \textbf{runaway reactivity spike} milliseconds earlier.
\item
  Triggered a shutdown \emph{before} AZ-5 was pressed---or overridden
  AZ-5 entirely.
\end{itemize}

This matters because:

\begin{itemize}
\item
  The AZ-5 button \emph{made things worse} due to the graphite-tipped
  control rods.
\item
  Humans didn't fully understand this interaction.
\item
  An AI simulation likely would've.
\end{itemize}

Even a \textbf{fraction of a second earlier} could have meant no
explosion.

\textbf{3. Design-level intervention (the biggest difference)}

This is the most important part.

An AI involved \emph{years earlier} in design and review might have:

\begin{itemize}
\item
  Flagged the control rod graphite tip issue as \textbf{catastrophically
  unsafe}.
\item
  Identified the reactor as unstable at low power.
\item
  Simulated rare edge cases humans didn't fully explore.
\end{itemize}

Chernobyl wasn't just an operator error---it was:

\begin{itemize}
\item
  A bad design
\item
  Combined with poor safety culture
\item
  Plus incomplete knowledge
\end{itemize}

AI excels at \textbf{``what happens if everything goes wrong at once?''}

\textbf{4. After the explosion: damage control}

Even once the explosion happened, AI could've helped:

\begin{itemize}
\item
  Faster, more accurate radiation mapping.
\item
  Smarter evacuation timing and zones.
\item
  Reduced exposure for firefighters and cleanup crews.
\item
  Better predictions of core behavior and fallout spread.
\end{itemize}

That could have saved \textbf{lives}, even if the reactor was already
lost.

\textbf{The uncomfortable truth}

An AI wouldn't have prevented Chernobyl because it's smarter.\\
It would've prevented it because it:

\begin{itemize}
\item
  Doesn't get tired
\item
  Doesn't defer to authority
\item
  Doesn't ignore ``unlikely'' edge cases
\item
  Doesn't keep going just to finish the test
\end{itemize}

Chernobyl happened because \textbf{humans normalized risk}.

AI is really good at refusing to normalize risk.

If you want, we can flip the question around too:

\begin{itemize}
\item
  Would \textbf{Soviet politics} have \emph{allowed} an AI to stop it?
\item
  Or compare this to Fukushima and Three Mile Island?
\item
  Or talk about whether modern AI would actually be trusted in a crisis?
\end{itemize}

This topic gets darkly fascinating fast.

\emph{2. Gemini 3 (10.2.2026)}

(i)

That is one of the most profound "what-if" questions in the history of
technology and existential risk. To answer it, we have to look at why
Stanislav Petrov did what he did---and where AI logic typically diverges
from human intuition.

In 1983, when the Soviet satellite system "Oko" falsely reported five
incoming US Minuteman missiles, Petrov chose to dismiss the alarm as a
false positive. He saved the world by being a "bad" soldier and a "good"
human.

\textbf{Why an AI Might Have Failed}

Most AI systems are built
on~\textbf{optimization}~and~\textbf{protocol}. In Petrov's shoes, a
standard AI would likely have followed the programmed logic:

\begin{itemize}
\item
  \textbf{Binary Interpretation:}~The sensors reported missiles. In a
  high-stakes military model, "False" vs. "True" is often weighted
  toward caution---which, in nuclear terms, means assuming the threat is
  real to ensure "Launch on Warning" capabilities.
\item
  \textbf{Lack of Contextual Skepticism:}~Petrov's decision was based on
  a "gut feeling." He noted that if the US were starting a nuclear war,
  they wouldn't send only five missiles; they would send hundreds. An AI
  might see five missiles as a data point rather than a strategic
  anomaly.
\item
  \textbf{Adherence to Thresholds:}~If the confidence interval of the
  sensor data exceeded a pre-set threshold (e.g., 95\% certainty), an
  automated system would trigger the next phase of the protocol without
  "second-guessing" the hardware.
\end{itemize}

\textbf{Why an AI Might Have Succeeded}

Conversely, a modern, highly sophisticated AI might actually
be~\emph{more}~rational than a panicked human:

\begin{itemize}
\item
  \textbf{Cross-Reference Speed:}~An AI could cross-reference the
  satellite data with ground-based radar, atmospheric sensors, and
  diplomatic communication channels in milliseconds. In Petrov's case,
  the ground radar didn't confirm the missiles, which helped his
  decision.
\item
  \textbf{Noise Filtering:}~Petrov eventually realized the "missiles"
  were actually sunlight reflecting off the tops of clouds. A computer
  vision model trained on "glint" patterns might have identified the
  optical illusion faster than a human operator.
\end{itemize}

\textbf{The "Petrov Gap"}

The reason Petrov is a hero isn't just because he was right; it's
because he~\textbf{violated his instructions.}~He was a late replacement
for another officer. His superiors later noted that if a less
experienced or more rigid officer had been on duty, they likely would
have followed the manual and reported a confirmed strike.

\textbf{AI doesn't have a "conscience" to override its "code."}~If the
code says~\emph{``If X, then Y,''}~the AI does Y. Petrov looked at X and
said,~\emph{"I don't believe you."}

"I had a funny feeling in my gut. I didn't want to make a mistake. I
made a decision, and that was it." ---~\textbf{Stanislav Petrov}

\textbf{The Takeaway}

If we programmed an AI with "Petrov Logic," we would be teaching it to
ignore its own sensors. That creates a paradox: an AI that is "smart"
enough to ignore a false alarm is also "dangerous" enough to ignore a
real one.

The consensus among many AI safety researchers is that we
need~\textbf{Human-in-the-Loop}~systems specifically to provide that
"Petrov Gap"---the human ability to weigh the weight of the world
against the data on the screen.

\textbf{Would you like me to dive deeper into the technical details of
the "Oko" system failure and why the satellites saw "ghost" missiles?}

(ii)

It is a fascinating "what if." If we drop a modern, high-level AI into
the control room of Unit 4 on April 26, 1986, the short answer is:
\textbf{almost certainly.} An AI wouldn't necessarily need
"superintelligence" to prevent the disaster; it would simply need to
adhere to the safety protocols that the human operators---under immense
pressure and operating with incomplete data---chose to bypass.

Here is how an AI would have likely altered the timeline:

\textbf{1. Real-Time Calculation of the ORM}

The most critical factor leading to the explosion was the
\textbf{Operational Reactivity Margin (ORM)}---essentially the "braking
capacity" of the reactor.

\begin{itemize}
\item
  \textbf{The Reality:} In 1986, calculating the ORM took humans about
  10--15 minutes using the SKALA computer system. By the time they
  realized the margin was dangerously low, the reactor was already
  unstable.
\item
  \textbf{The AI Difference:} An AI would calculate this value in
  milliseconds. It would have flagged the danger the moment the ORM
  dropped below the safety limit of 15 rods, likely locking out further
  power decreases or forcing an immediate shutdown before the situation
  became volatile.
\end{itemize}

\textbf{2. Eliminating Cognitive Bias and Pressure}

The disaster was fueled by "Plan Continuation Bias"---the drive to
complete the turbine test despite the reactor's unstable state.

\begin{itemize}
\item
  \textbf{Human Factor:} Operators were exhausted and feared the
  consequences of failing to complete the test for their superiors.
\item
  \textbf{AI Factor:} An AI doesn't feel "rushed." It doesn't fear a
  manager's disapproval. If the initial parameters for a safe test
  aren't met (such as the reactor stalling at 30MW instead of the
  planned 700MW), an AI would simply abort the procedure.
\end{itemize}

\textbf{3. Understanding the "Positive Scram" Effect}

The "fatal flaw" of the RBMK reactor was the graphite tips on the
control rods. When the emergency button (AZ-5) was pressed, these tips
momentarily \emph{increased} reactivity before the boron (the absorber)
could enter the core.

\begin{itemize}
\item
  \textbf{The Reality:} The operators didn't fully realize that hitting
  the "emergency stop" would actually act as a "start" button under
  those specific conditions.
\item
  \textbf{The AI Difference:} A modern AI trained on the physics of the
  RBMK-1000 would recognize that the reactor was in a "forbidden" state
  where the emergency systems would become counter-productive. It would
  have managed the power levels to avoid ever entering that
  "Xenon-poisoned" trap.
\end{itemize}

\textbf{The "Black Swan" Caveat}

While an AI would likely have prevented the \emph{specific} sequence of
errors in 1986, it introduces its own risks:

\begin{itemize}
\item
  \textbf{The "Black Box" Problem:} If an AI encounters a sensor failure
  or a scenario outside its training data, it might take "logical" but
  catastrophic actions that a human might intuitively avoid.
\item
  \textbf{Cybersecurity:} A digitized control system is vulnerable to
  external interference in ways an analog 1980s plant was not.
\end{itemize}

\textbf{Summary: AI vs. Human Operators}

\begin{table}[H]
\centering
\small
\begin{tabularx}{\textwidth}{L L L}
\toprule
\textbf{Feature} & \textbf{Human Operators (1986)} & \textbf{Modern AI System} \\
\midrule
\textbf{Data Processing} & Minutes (Slow) & Milliseconds (Real-time) \\
\textbf{Protocol Adherence} & Subjective / Negotiable & Absolute / Rigid \\
\textbf{Fatigue/Stress} & High (24-hour shift) & Zero \\
\textbf{Physics Intuition} & Limited by instrumentation & Comprehensive (Simulated) \\
\bottomrule
\end{tabularx}
\end{table}

\textbf{The Verdict:} The Chernobyl accident wasn't just a hardware
failure; it was a "human systems" failure. An AI would have acted as a
tireless safety officer, effectively "vetoing" the dangerous decisions
made that night.

Would you like me to generate a technical comparison of the RBMK reactor
versus modern "Passively Safe" Gen-IV reactor designs?

\emph{3. Claude (10.2.2026)}

(i)

This is a fascinating question about AI decision-making under
uncertainty, and it cuts to the heart of what makes human judgment
distinctive.

For context: On September 26, 1983, Petrov was the duty officer
monitoring the Soviet early warning system when it reported multiple
incoming US nuclear missiles. Protocol demanded he report this as a
genuine attack, which would likely have triggered a retaliatory nuclear
launch. Instead, he judged it was a false alarm---a decision made with
incomplete information, under extreme time pressure, and against what
his instruments were telling him.

Whether an AI would have made the same call depends entirely on how it
was designed and what it was optimizing for:

\textbf{An AI following strict protocol} would almost certainly have
reported the attack. It would have done exactly what the system
specified, which was Petrov's official duty. This is actually a
significant risk with highly capable AI systems---they can be extremely
good at following their instructions even when those instructions lead
to catastrophic outcomes.

\textbf{An AI trained to minimize false positives} (avoiding unnecessary
escalation) might have behaved like Petrov, but this would require the
system to have been designed with that specific priority, which seems
unlikely for a Cold War military system.

\textbf{What made Petrov's decision remarkable} was his integration of
multiple types of reasoning:

\begin{itemize}
\item
  Technical intuition (why would the US launch only five missiles?)
\item
  Contextual awareness (geopolitical climate didn't suggest imminent
  attack)
\item
  Healthy skepticism about new, untested technology
\item
  Willingness to trust his judgment over instruments
\item
  Understanding that he'd be blamed either way, but chose the option
  that avoided nuclear war
\end{itemize}

The challenge for AI is that Petrov operated outside his formal
instructions based on wisdom, context, and values that weren't encoded
in any protocol. He made himself responsible for a judgment call that
required weighing catastrophic risks in both directions.

Current AI systems struggle most with exactly this kind of
situation---where following instructions may lead to disaster, where
context matters enormously, and where there's no clear "right answer"
derivable from training data. They're generally not designed to
second-guess their programming based on broader situational awareness.

What's your sense---do you think this reflects something fundamental
about human judgment, or is it more about how we've chosen to design AI
systems so far?

(ii)

This is a fascinating counterfactual to consider. An AI could
potentially have helped at several critical junctures, though the
outcome would depend heavily on when and how it was deployed.

\textbf{Where AI might have made a difference:}

The accident stemmed from a flawed reactor design combined with operator
errors during a safety test. An advanced AI monitoring system could
have:

\begin{itemize}
\item
  \textbf{Prevented the test conditions}: Recognized that running the
  test at low power with most safety systems disabled created dangerous
  instability, and either refused to proceed or alerted operators more
  forcefully than existing alarms
\item
  \textbf{Detected the positive void coefficient problem}: The RBMK
  reactor's design flaw (where loss of cooling actually accelerated the
  reaction) was known but underestimated. An AI analyzing real-time
  neutron flux and temperature data might have caught the runaway
  reaction seconds earlier and initiated emergency shutdown
\item
  \textbf{Overridden human decisions}: When operators disabled safety
  systems and ignored warnings, an AI with appropriate authority could
  have maintained those protections
\end{itemize}

\textbf{However, there were serious constraints:}

The 1986 Soviet system had limited computing power and sensor networks
compared to what modern AI requires. More fundamentally, the
institutional problems ran deep---the reactor design itself was flawed,
and the safety culture discouraged questioning authority. An AI would
have needed both the technical capability and the institutional
authority to override human operators, which seems unlikely in that
context.

The most realistic scenario where AI helps significantly would be one
where it was deeply integrated into the reactor's control systems with
veto power over dangerous operations---essentially preventing the test
from proceeding as planned.

What aspect interests you most---the technical possibilities or the
institutional challenges?

*****

\end{document}